\pdfoutput=1
\PassOptionsToPackage{dvipsnames,svgnames,table}{xcolor}

\documentclass[11pt]{article}

\usepackage[]{ACL2023}

\usepackage{times}
\usepackage{latexsym}

\usepackage[T1]{fontenc}

\usepackage[utf8]{inputenc}

\usepackage{microtype}

\usepackage{inconsolata}

\usepackage{graphicx} 
\usepackage[table, dvipsnames]{xcolor}
\usepackage{amsmath, amssymb}
\usepackage{mathtools}
\usepackage{longtable}
\usepackage{float}
\usepackage{booktabs}    
\usepackage{multirow}

\title{Learning from Committee: Reasoning Distillation from a Mixture of Teachers with Peer-Review}

\author{
 \textbf{Zhuochun Li\textsuperscript{1}},
 \textbf{Yuelyu Ji\textsuperscript{1}},
 \textbf{Rui Meng\textsuperscript{2}},
 \textbf{Daqing He\textsuperscript{1}},
\\
 \textsuperscript{1}School of Computing and Information, University of Pittsburgh, Pittsburgh, USA \\
 \textsuperscript{2}Salesforce Research
\\
   \{zhl163, yuj49, dah44\}@pitt.edu, memray0@gmail.com
}

\begin{document}
\maketitle
\begin{abstract}
While reasoning capabilities typically emerge in large language models (LLMs) with tens of billions of parameters, recent research focuses on improving smaller open-source models through knowledge distillation (KD) from commercial LLMs. However, many of these studies rely solely on responses from a single LLM as the gold rationale, unlike the natural human learning process, which involves understanding both the correct answers and the reasons behind mistakes. In this paper, we introduce a novel \textbf{F}ault-\textbf{A}ware Dist\textbf{I}llation via Peer-\textbf{R}eview (\textbf{FAIR}) approach: 1) instead of merely obtaining rationales from teachers, our method asks teachers to identify and explain the student's mistakes, providing customized instruction learning data; 2) we design a simulated peer-review process between teacher LLMs, and selects only the generated rationales above the acceptance threshold, which reduces the chance of teachers guessing correctly with flawed rationale, improving instructional data quality. Comprehensive experiments and analysis on mathematical, commonsense, and logical reasoning tasks demonstrate the effectiveness of our method. Our code is available at \url{https://github.com/zhuochunli/Learn-from-Committee}.
\end{abstract}

\section{Introduction}
Large Language Models (LLMs) have proven to be highly effective in addressing a wide range of complex tasks~\cite{ni2024earnings, fan2024towards}, including mathematical reasoning~\cite{lewkowycz2022solving, imani2023mathprompter}, commonsense reasoning~\cite{zhao2024large, achiam2023gpt}, and logical reasoning~\cite{liu2023evaluating, xu2023large}. However, these emerging reasoning abilities tend to manifest only in LLMs with more than 100 billion parameters, while smaller models struggle to exhibit such capabilities~\cite{wei2022emergent}. Despite this, related research~\cite{touvron2023llama, zeng2022glm} has shown that smaller language models, particularly those with fewer than 10 billion parameters, can perform similarly to larger models in terms of following human instructions. However, it is challenging to prompt smaller Language Models (LMs) to generate reasoning steps by Chain-of-Thought (CoT) prompts~\cite{wang2023democratizing}. Moreover, most existing reasoning datasets lack high-quality rationale~\cite{gurrapu2023rationalization} due to the high cost of manual annotations.

\begin{figure}[tb]
    \centering
     \includegraphics[width=0.5\textwidth,height=0.22\textwidth]{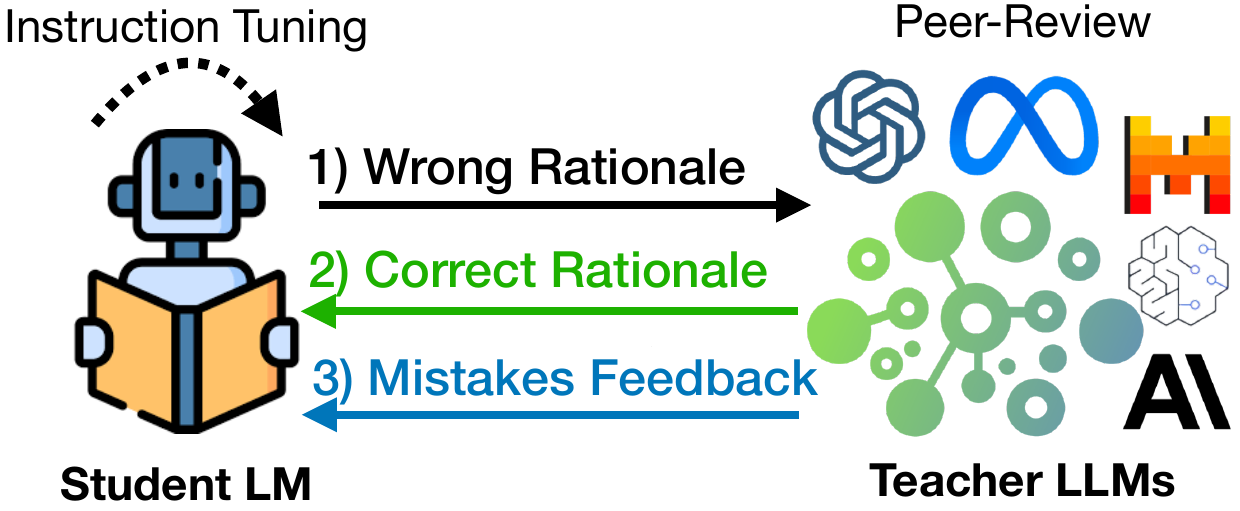}
     \caption{Student LM learns from multiple teacher LLMs via Peer-Review distillation.}
    \label{fig_simple}
    \vspace{-0.45cm}   
\end{figure}

To address these challenges, distilling the capabilities of LLMs emerges as a resource-friendly and effective strategy. DeepSeek-R1~\cite{guo2025deepseek} demonstrates that distilling reasoning patterns from larger models can outperform RL-derived patterns on smaller models. Through collecting rationales generated by LLMs for instruction tuning, previous studies have been able to distill the private LLMs' reasoning abilities into smaller models~\cite{wang2022self, ho2023large, magister2022teaching, fu2023specializing}. However, most of these efforts fall within the scope of Labeling Knowledge Distillation~\cite{xu2024survey}, where LLMs are primarily used to annotate data for training smaller models, without utilizing smaller model's outputs as feedback to generate customized instruction data to improve the LM in return. As a result, LLMs remain unaware of the limitations of smaller models.

Furthermore, prior research typically employs only one LLM as the teacher, which can introduce more biased training data compared to using multiple teacher LLMs during distillation. Therefore, we propose using multiple LLMs from different organizations as teachers to provide more impartial and diverse training data. Additionally, we design a simulated peer-review process between teacher LLMs, where the rationale generated by one LLM is reviewed by other LLMs. Only the rationales that pass this peer-review process are included in the training dataset. This method reduces the likelihood of flawed rationales, even when a correct answer is provided, thereby, improving the overall quality of the training data for instruction tuning.

To this end, we propose a Fault-Aware Distillation via Peer-Review (FAIR) knowledge distillation method from multiple LLMs, as briefly shown in Figure~\ref{fig_simple}. Inspired by the natural human learning process~\cite{konold2004using}, we argue that students should not only know \textbf{what} is the correct answer but also learn \textbf{why} they made mistakes. Therefore, in addition to providing the correct rationale generated by the teacher LLMs, we also present the student model’s mistakes to the teacher LLMs and return the mistake-specific feedback. Furthermore, inspired by the multi-agent evaluation framework of~\citet{nan2023evaluating}, we employ multiple LLMs as teachers. Each teacher LLM’s answer is reviewed by the other teachers, and only the responses that pass this peer-review process are included in the instruction training dataset. We believe that this peer-review mechanism can significantly reduce biased or flawed rationales, leading to improved distillation performance. In summary, the contributions of our work are as follows:
\vspace{-5pt}
\begin{enumerate}
    \item The \textbf{F}ault-\textbf{A}ware Dist\textbf{I}llation via Peer-\textbf{R}eview (\textbf{FAIR}) approach is introduced to help student LM learn form not only the correct rationale but also the feedback on their own mistakes provided by teacher LLMs, which builds a comprehensive instruction tuning method aimed at enhancing the student LM’s general reasoning abilities.
    \item We design a simulated Peer-Review mechanism between teacher LLMs to filter out flawed rationales and improve the confidence of instruction tuning data.
    \item Our work provides a \textbf{comprehensive benchmark} on the mathematical, commonsense, and logical reasoning tasks. Experiments and comparisons with concurrent works demonstrate the effectiveness of our method in distilling the reasoning ability of teacher LLMs.
\end{enumerate}

\section{Related Work}
\textbf{LLM Reasoning} \ \ Recent studies focus on provoking the thought processes of LLMs, validating their effectiveness in reasoning tasks~\cite{wei2022chain, imani2023mathprompter, fu2023specializing}. Various techniques have been developed to enhance LLM reasoning abilities~\cite{chu2023survey, xu2024can, chen2023mcc, li2024mosaic}. Chain-of-Thought (CoT)~\cite{wei2022chain} improves reasoning by prompting LLMs to generate intermediate natural language thought processes. ~\citet{huang2022large} demonstrates that LLMs can self-improve through self-training on majority voting data. ~\citet{chung2024scaling} showed that smaller LMs can acquire CoT skills by training on rationales. The work s1~\cite{muennighoff2025s1} proves the significance of high-quality CoT data on model performance. In this paper, we further show that the CoT performance of smaller LMs can be improved through integrated instruction learning using CoT data selected by LLMs via peer-review. \\
\textbf{Knowledge Distillation from LLMs} \ \ Distilling knowledge from LLMs by fine-tuning smaller language models using high-quality data collected from LLMs has become a prominent research direction~\cite{xu2023baize, li2024turning, guo2025deepseek}. This approach serves as an effective method for transferring the emergent abilities of black-box LLMs to smaller open-source models. However, while recent works~\cite{ho2023large, shridhar2022distilling, guo2024exploring} use LLM-generated reasoning rationales as supervisory signals, they often overlook providing student models with feedback on their mistakes when their answers are incorrect. To address this, we collect both the correct rationale and mistake-specific feedback~\cite{jiang2023lion} for student models' wrong answers from LLMs, integrating them into instruction tuning to enhance the overall reasoning capabilities of the student models. Moreover, unlike previous studies that depend on a single teacher LLM~\cite{chenglin2023mixed, zhu2024pad} or intermediate roles such as mentors~\cite{lee2024mentor} and teaching assistant (TA)~\cite{zhou2024teaching}, we employ multiple LLMs~\cite{tian2024tinyllm, sun2023corex} as teachers to increase the diversity of generated data. Finally, compared to peer-review methods in LLMs for evaluation~\cite{ning2024peer, chu2024pre}, we design a simulated peer-review process to ensure high-quality instruction training data, thereby improving the distillation performance.
\begin{figure*}[htb]   
    \centering
     \includegraphics[width=1\textwidth,height=0.44\textwidth]{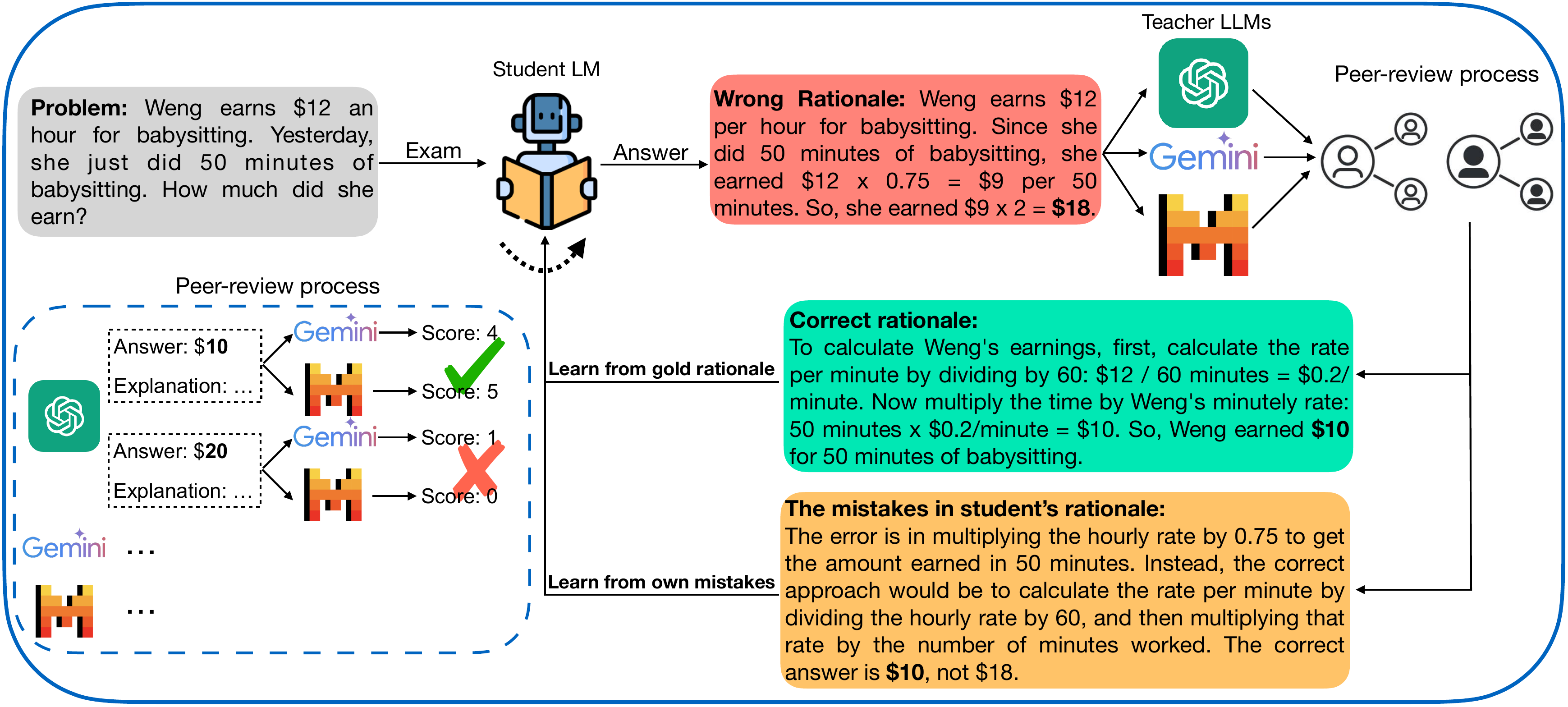}
     \caption{Overview of our \textbf{F}ault-\textbf{A}ware Dist\textbf{I}llation via Peer-\textbf{R}eview (\textbf{FAIR}) method. The specific structure of the peer-review process, which is used to generate the correct rationale, is explained in the bottom-left sub-figure.}
    \label{fig_overview}
    \vspace{-0.5cm}
\end{figure*}

\section{Method}
As illustrated in Figure~\ref{fig_overview}, we introduce a \textbf{F}ault-\textbf{A}ware Dist\textbf{I}llation via Peer-\textbf{R}eview (\textbf{FAIR}) knowledge distillation method that empowers the student model to improve by learning from its own mistakes and the correct answers generated by multiple teacher models. Specifically, our instruction learning procedure involves four major steps: (1) the student LM takes an ``exam`` on the training set to identify mistakes that are incorrectly generated rationales; (2) we then craft various prompts that incorporate the question and the student’s wrong rationale to prompt the teacher LLMs to generate correct answers and provide feedback on the student's errors respectively; (3) a simulated peer-review process is conducted among the teacher LLMs to produce highly confident instructional data; (4) finally, the student model learns to reason through instruction learning based on the peer-reviewed correct answers and tailored corrections on its mistakes provided by the teacher LLMs.

\subsection{Collecting Mistakes on Student Model}
We aim to gather samples from reasoning benchmarks in which the student model incorrectly answers questions. These samples will be used to create customized instructional data from the teacher models. To achieve this, the student model undergoes an “exam” on the training set \emph{$D_{train}$} to assess its reasoning ability and collect the mistake set \emph{$D_{mistake}$}, which are the samples containing incorrect rationales and answers. Specifically, given a dataset \emph{$D = \{x, y\}$}, where \emph{$x$} is the question and \emph{$y$} is the gold answer, we propose to input the question \emph{$x$} into the student model \emph{$f$} to generate the output \emph{$f(x) = [r', y']$}. Here, the square brackets denote the concatenation of the student model's rationale \emph{$r'$} and answer \emph{$y'$}, with the answer typically at the end of the output. Since the correct rationale \emph{$r$} is often not provided in \emph{$D_{train}$}, we follow~\citet{wang2023democratizing}'s work by considering \emph{$r'$} as the wrong rationale if \emph{$y' \neq y$}. Finally, the mistake set \emph{$D_{mistake}$} is collected as follows:
\begin{equation}
\label{eq:1}
\resizebox{.89\hsize}{!}{$
D_{mistake}=\{(x,r',y') \ | \ (x,y) \in D_{train}, \ y' \neq y\}
$}
\end{equation}
where \emph{$x$} is the question, \emph{$r'$} is the wrong rationale, \emph{$y$} and \emph{$y'$} are correct and wrong final answer.

The collected mistake set \emph{$D_{mistake}$} highlights the student’s reasoning weaknesses and will be utilized for the following purposes:
\begin{itemize}
\vspace{-5pt}
\item [1)] 
  Providing the incorrectly answered questions for the teacher LLMs to generate correct rationales.
  \vspace{-5pt}
  \item [2)]
   Using the student's incorrect rationales to prompt the teacher LLMs to identify errors and create customized mistakes feedback.
\end{itemize}
\begin{figure}[htb]
    \centering
     \includegraphics[width=0.37\textwidth,height=0.35\textwidth]{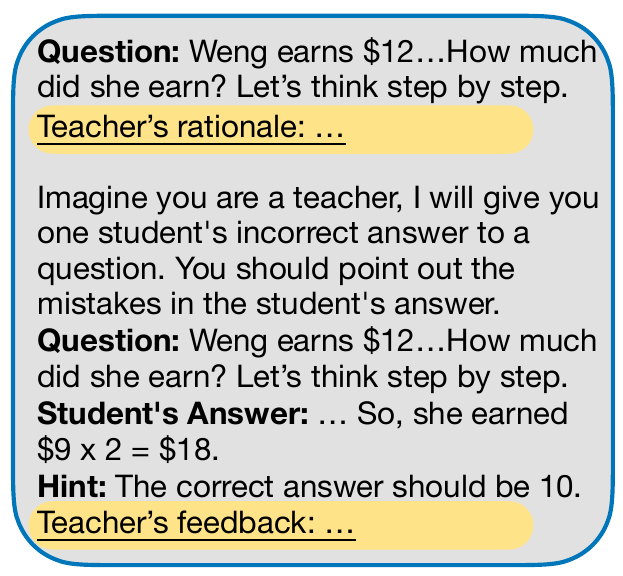}
     \caption{The prompt template \emph{$P_{rt}$} (first) and \emph{$P_{fb}$} (second) for asking teacher LLMs to generate rationale and mistakes feedback. The part colored in yellow is the teacher's output.}
    \label{fig_prompt_rf}
    \vspace{-0.5cm}
\end{figure}

\subsection{Inquiring Teacher LLMs with Student's Mistakes}
We expect that the teacher LLM should act as a reasoning instructor who can identify student's mistakes and provide tailored feedback rather than merely as an answer provider. Therefore, we query the teacher LLMs with the student's incorrectly answered questions, aiming for them to generate the correct rationale and identify specific errors in the student's mistakes. We believe that customized training data, which includes both ``what'' the correct answer is and ``why'' the mistakes were made, can effectively address the student’s weaknesses. For prompt \emph{$P_{fb}$} to gather feedback on the student's mistakes, we follow~\citet{zelikman2022star} by adding a hint that explicitly provides the correct answer to the question, ensuring more accurate responses. The detailed prompt templates are shown in Figure~\ref{fig_prompt_rf}. In detail, for each sample \emph{$(x, r', y') \in D_{mistake}$}, we request each teacher \emph{$\mathcal{M}_{T}^{k}$} from the total of N teacher LLMs to generate its own feedback \emph{$f_{k}$}, which will be collected as the mistakes feedback set \emph{$D_{feedback}$}:
\begin{equation}
\label{eq:2}
\resizebox{1\hsize}{!}{$
\begin{aligned}  
&f_{k}=\mathcal{M}_{T}^{k}(P_{fd}(x, r', y)) \\
&D_{feedback}=\{(x, r', f_{k}) \ | \ (x, r', y') \in D_{mistake}, 1 \leq k \leq N\} \\
\end{aligned}
$}
\end{equation}
where \emph{$\mathcal{M}_{T}^{k}(x)$} represents the k-th teacher LLM’s output when given x as the input. \emph{$P_{fb}(x)$} denotes the prompt template filled in with x to generate mistakes feedback.

\subsection{Simulating Peer-Review Between Teacher Models}
During our experiments, we observe that the rationales provided by teacher LLMs are not always accurate, even when the final answer matches the gold answer. This discrepancy is rare in mathematical tasks, where there is often a strict correlation between the correctness of the rationale and the final answer number because of the inherent nature of mathematics. However, for multiple-choice questions, such as those in the commonsense StrategyQA~\cite{geva2021didaristotleuselaptop} (True or False) and logic LogiQA~\cite{liu2020logiqachallengedatasetmachine} (A, B, C, D) benchmarks, there are instances where a correct rationale may lead to an incorrect final choice, or a wrong rationale might result in a correct final choice. See Appendix~\ref{sec:appendix_pr} for more peer-review examples on different benchmarks.
\begin{figure}[tb]   
    \centering
     \includegraphics[width=0.4\textwidth,height=0.38\textwidth]{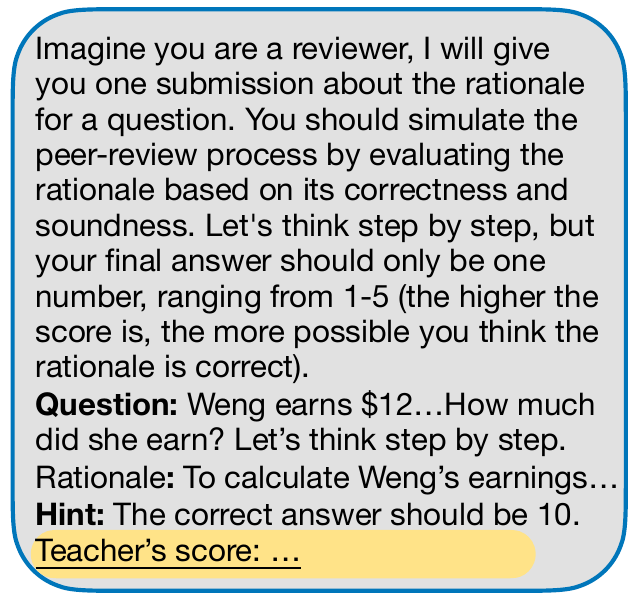}
     \caption{The prompt template \emph{$P_{pr}$} for asking teacher LLMs to perform peer-review process. The part colored in yellow is the teacher's output.}
    \label{fig_prompt_pr}
    \vspace{-0.5cm}
\end{figure}

To address this issue and avoid having teacher LLMs ``\textbf{guess}'' the correct answer without well-grounded reasoning steps, we propose a simulated peer-review process among teacher LLMs. Since most relevant datasets do not provide gold rationales, we assume that each LLM's rationale should be reviewed and scored by peer LLMs, which is inspired by the multi-agent evaluation framework of~\citet{nan2023evaluating}. Only those rationales that pass this peer-review process with high confidence will be included in the final instructional tuning dataset. Figure~\ref{fig_overview} has explained the peer-review process. For the rationale generated by each teacher LLM, we incorporate it into the designed peer-review prompt \emph{$P_{pr}$} shown in Figure~\ref{fig_prompt_pr} and request all other LLMs to score it. Specifically, assume we have N different teacher LLMs \emph{$\mathcal{M}_{T}^{1}, \mathcal{M}_{T}^{2},...,\mathcal{M}_{T}^{N}$}. For the k-th teacher LLM \emph{$\mathcal{M}_{T}^{k}$}, we obtain its generated rationale \emph{$r_{k}$} by:
\begin{equation}
\label{eq:3}
r_{k}=\mathcal{M}_{T}^{k}(P_{rt}(x))    
\end{equation}
where \emph{$\mathcal{M}_{T}^{k}(x)$} represents the k-th teacher LLM’s output when given x as the input. \emph{$P_{rt}(x)$} denotes the rationale prompt template filled in with x.

Subsequently, we ask each teacher except \emph{$M_{T}^{k}$} to peer-review this rationale \emph{$r_{k}$} and score it. The scores are collected to form the score set \emph{$Score(r_k)$} for rationale \emph{$r_{k}$}. Only the rationale \emph{$r_{k}$} with an average score \emph{$Avg(Score(r_{k}))$} exceeding the acceptance threshold \emph{$Th$} will be included in the rationale set \emph{$D_{rationale}$}: 
\begin{equation}
\label{eq:4}
\resizebox{1\hsize}{!}{$
\begin{aligned}  
Score(r_k) &= \{\mathcal{M}_{T}^{i}(P_{pr}(x, r_{k}, y)) \ | \ 1 \leq i \leq N \ and \ i \neq k \} \\
D_{rationale} &= \{(x,r_{k}) \ | \ if \ Avg(Score(r_{k})) \geq Th, 1 \leq k \leq N \} \\
\end{aligned}
$}
\end{equation}
where \emph{$\mathcal{M}_{T}^{i}(x)$} represents the i-th teacher LLM’s output with input x. \emph{$P_{pr}(x)$} denotes the peer-review prompt template filled in with x to generate score.


\subsection{Instruction Tuning for Student Models}
The reasoning ability of the student LM can be enhanced through instruction tuning~\cite{wei2021finetuned}, which incorporates both verified rationales and customized mistake corrections provided by the teacher models. See Appendix~\ref{sec:appendixB} for explicit instruction tuning templates on different benchmarks.\\
\textbf{Learning from Teacher's Rationales} \ \ The rationales generated by the teacher LLMs are specifically tailored to address the student’s weaknesses, identified through the student’s previous exam. According to Equation~\ref{eq:4}, these collected rationales are combined into the set \emph{$D_{rationale}$} as the correct rationales, which are then used to fine-tune the student LM. For the instruction tuning process, we aim for the student model, when given the question \emph{$x$} as the instruction, to produce an answer that closely aligns with the corresponding rationale \emph{$r$} in \emph{$D_{rationale}$}. The loss function for learning from the teacher's rationale is defined as follows:
\begin{equation}
\label{eq:5}
\mathcal{L}_{\text{rationale}} = \mathbb{CE}(\mathcal{M}_S(x), r), \ for \ r \in D_{\text{rationale}}
\end{equation}
where \emph{$\mathbb{CE}$} denotes the Cross-Entropy function, and \emph{$\mathcal{M}_S(x)$} represents the student LM's output when given \emph{$x$} as the input.\\
\textbf{Learning from Student's Mistakes} \ \ In addition to learning from correct rationales, we propose that the student model should also learn from its own mistakes, simulating the typical human learning process. This approach helps the student not only grasp the correct answers but also understand the reasons behind the errors. To facilitate this, we constructed the feedback set \emph{$D_{feedback}$}, based on Equation~\ref{eq:2}, which provides feedback on the student's mistakes. Through this process, we expect the student LM to learn the teacher's reasoning capabilities and generate outputs that closely align with the teacher's feedback \emph{$f$} when given instructions to identify its own mistakes. Finally, the loss function for learning from mistakes feedback is defined as follows:
\begin{equation}
\label{eq:6}
\resizebox{.89\hsize}{!}{$
\mathcal{L}_{\text{feedback}} = \mathbb{CE}\left(\mathcal{M}_S(x \oplus r'), f \right), \ for \ f \in D_{feedback}
$}
\end{equation}
where \emph{$\mathbb{CE}$} denotes the Cross-Entropy function, and \emph{$\oplus$} represent the string concatenation. \emph{$\mathcal{M}_S(x \oplus r')$} represents the student LM's output when given \emph{$x \oplus r'$} as the input. \\
\textbf{Joint Learning} \ \ The final optimization process integrates learning from both correct answers and the teachers' customized mistakes feedback. Therefore, the instruction learning losses from Equation~\ref{eq:5} and Equation~\ref{eq:6} are combined as follows:
\begin{equation}
\label{eq:7}
\mathcal{L}=\alpha \cdot \mathcal{L}_{\text{feedback}} + (1-\alpha) \cdot \mathcal{L}_{\text{rationale}}
\end{equation}
where \emph{$\alpha$} controls the impact of learning from mistakes, balancing the two learning objectives.

\section{Experiments}
\subsection{Datasets}
We focus on evaluating reasoning abilities with various datasets, including mathematical reasoning with GSM8K~\cite{cobbe2021training} and SVAMP~\cite{patel2021nlp}, commonsense reasoning with StrategyQA~\cite{geva2021didaristotleuselaptop}, and logical reasoning with LogiQA~\cite{liu2020logiqachallengedatasetmachine}. All datasets were downloaded from Huggingface, utilizing the standard train/test set split. Datasets statistics are shown in Appendix~\ref{sec:appendixA.1}.

\begin{table*}[ht]
\centering
\resizebox{2.1\columnwidth}{!}{\begin{tabular}{l|cl|cccc}
\toprule 
\multicolumn{1}{c|}{\multirow{2}{*}{\centering \textbf{Method}}}  & 
\multicolumn{1}{c}{\multirow{2}{*}{\centering \textbf{\# Params}}} &
\multicolumn{1}{c|}{\multirow{2}{*}{\centering \textbf{Distillation Teachers}}} & 
\multicolumn{2}{c}{\textbf{Mathematical}} & \textbf{Commonsense} & \textbf{Logical} \\
 & & & GSM8K & SVAMP & StrategyQA & LogiQA \\ \hline
\rowcolor{lightgray} \multicolumn{7}{l}{\textbf{Teacher LLMs}} \\ 
GPT-3.5-Turbo & 175B & - & 78.01* & 82.30* & 70.92* & 40.55* \\ 
Gemini-1.0-Pro & - & - & 76.42* & 81.10* & 67.03* & 39.94 \\ 
Mixtral-8x7B-Instruct-v0.1 & 46.7B & - & 74.40* & 81.60* & 72.83* & 34.19* \\ \hline
\rowcolor{lightgray} \multicolumn{7}{l}{\textbf{Student LM Baselines}} \\ 
CodeT5-Large+PaD~\citep{zhu2024pad} & 770M & GPT-3.5-Turbo & 44.90* & 51.00* & - & - \\
Qwen2-1.5B+SIKeD~\citep{adarsh2025siked} & 1.5B & Llama3-70B & 64.97* & 75.40* & - & - \\
GPT-J+Self-Reflection~\citep{wang2023democratizing} & 6B & ChatGPT & 33.10* & 55.00* & 65.90* & - \\ 
Llama-7B+NCE~\citep{li2024turning} & 7B & GPT-3.5-Turbo, GPT-4 & 41.93* & 51.50* & - & - \\
Llama2-7B+ReversalMath~\citep{guo2024exploring} & 7B & GPT-4 & 52.10* & 59.20* & - & - \\
ORCA2-7B~\citep{mitra2023orca} & 7B & ChatGPT, GPT-4 & 47.23* & - & - & 35.02* \\ 
Llama3.1-8B+ReDistill~\citep{badri2025r1} & 8B & DeepSeek-R1 & 75.66* & 82.00 & - & - \\
T5-XXL+CoT~\citep{magister2022teaching} & 11B & PaLM, GPT-3 & 21.99* & - & 63.77* & - \\ \hline
\rowcolor{lightgray} \multicolumn{7}{l}{\textbf{Peer-Reviewed Distillation (Ours)}} \\ 
Llama2-7B-chat~\citep{touvron2023llama} & 7B & - & 15.62 & 39.67 & 47.02 & 18.74 \\
\hspace{8pt}+Teacher-Mixtral & 7B & Mixtral-8x7B-Instruct & 22.67 & 47.33 & 62.70 & 32.10 \\
\hspace{8pt}+Teacher-Gemini & 7B & Gemini-1.0-Pro & 26.84 & 49.33 & 57.93 & 32.72 \\
\hspace{8pt}+Teacher-GPT & 7B & GPT-3.5-Turbo & 30.71 & 51.67 & 60.12 & 31.04 \\
\hspace{8pt}+Teacher-Multiple, w/o Peer-Review & 7B & Multiple & 29.65 & 52.67 & 56.62 & 29.65 \\
\hspace{8pt}+Teacher-Multiple & 7B & Multiple & \textbf{36.24} & \textbf{59.50} & \textbf{67.69} & \textbf{36.25} \\ \hline
Qwen2.5-1.5B-Instruct~\citep{qwen2} & 1.5B & - & 64.44 & 77.00 & 53.86 & 19.97 \\
\hspace{8pt}+Teacher-Mixtral & 1.5B & Mixtral-8x7B-Instruct & 65.81 & 77.67 & 63.32 & 32.10 \\
\hspace{8pt}+Teacher-Gemini & 1.5B & Gemini-1.0-Pro & 66.26 & 78.67 & 60.41 & 33.95 \\
\hspace{8pt}+Teacher-GPT & 1.5B & GPT-3.5-Turbo & 68.01 & 79.33 & 62.45 & 34.25 \\
\hspace{8pt}+Teacher-Multiple, w/o Peer-Review & 1.5B & Multiple & 67.48 & 77.67 & 61.43 & 33.03 \\ 
\hspace{8pt}+Teacher-Multiple & 1.5B & Multiple & \textbf{72.48} & \textbf{81.00} & \textbf{68.12} & \textbf{38.71} \\ \hline
Llama3.1-8B-Instruct~\citep{dubey2024llama} & 8B & - & 74.00 & 81.67 & 63.03 & 36.56 \\
\hspace{8pt}+Teacher-Mixtral & 8B & Mixtral-8x7B-Instruct & 74.83 & 82.00 & 71.62 & 37.02 \\
\hspace{8pt}+Teacher-Gemini & 8B & Gemini-1.0-Pro & 76.42 & 82.33 & 66.96 & 39.94 \\
\hspace{8pt}+Teacher-GPT & 8B & GPT-3.5-Turbo & 77.94 & 83.00 & 70.16 & 40.86 \\
\hspace{8pt}+Teacher-Multiple, w/o Peer-Review & 8B & Multiple & 76.57 & 82.67 & 70.89 & 38.40 \\ 
\hspace{8pt}+Teacher-Multiple & 8B & Multiple & \textbf{79.30} & \textbf{84.33} & \textbf{73.07} & \textbf{43.16} \\ \bottomrule
\end{tabular}}
\caption{\label{table_results} Accuracy (\%) across various reasoning tasks with different distillation methods. * denotes the results are from the original paper or official document. ``Teacher-x'' indicates the specific teacher LLM used in the distillation experiment. The best performance among different student LMs in each benchmark is marked in \textbf{bold}.}
\vspace{-0.5cm}
\end{table*}

\subsection{Baselines}
To demonstrate the effectiveness of our method, we included the following baselines: (1) teacher LLMs and student LMs without fine-tuning, to show the impact of distilling reasoning abilities; (2) established distillation methods on smaller (CodeT5-Large+PaD~\citep{zhu2024pad}) and larger models (T5-XXL+CoT~\citep{magister2022teaching}); (3) GPT-J+Self-Reflection~\citep{wang2023democratizing}, from which we draw inspiration; (4) Qwen2-1.5B+SIKeD~\citep{adarsh2025siked}, for direct comparison on our Qwen2-1.5B; and (5) LLaMA-based models, including LLaMA-7B+NCE~\citep{li2024turning}, LLaMA2-7B+ReversalMath~\citep{guo2024exploring}, ORCA2-7B~\citep{mitra2023orca}, and LLaMA3.1-8B+ReDistill~\citep{badri2025r1}. As our work aims to benchmark diverse reasoning tasks, it is challenging to find comparable prior work covering all our datasets, so we compare against related methods on overlapping tasks. Furthermore, we exclude certain closely related works from our baseline comparisons either because they do not report results on any dataset overlapping with ours, or because their performance is inferior to that of the baselines already included.

\subsection{Implementation Details}
\textbf{Models} \ \ Since our work considers scenarios of limited resources, we intentionally selected entry-level teacher LLMs and smaller student models for distillation. We selected GPT-3.5-Turbo\footnote{\url{https://platform.openai.com/docs/models/gpt-3-5-turbo}}, Gemini-1.0-Pro~\cite{team2023gemini}, and Mixtral-8x7B-Instruct-v0.1~\cite{jiang2024mixtral} as the teacher LLMs. The selection motivations include the considerations of the expense and accessibility of the LLMs and their proved powerful NLP capabilities. \\
Among the three student models, we choose Llama2-7B-chat~\cite{touvron2023llama} as the backbone for its active community to compare performance, and Qwen2.5-1.5B-Instruct~\cite{qwen2} as well as Llama3.1-8B Instruct~\cite{dubey2024llama} to test the generalizability of FAIR method.  
The threshold in Equation~\ref{eq:4} was set to {$Th=4$} for high confident rationales. The parameter {$\alpha$} in Equation~\ref{eq:7} was set to {$\alpha=0.5$} to balance the impact of learning from mistakes. For data inference from teacher LLMs, we collect samples that have at least one peer-reviewed rationale and one feedback. During the training, we randomly select one feedback and one rationale for each sample. All evaluation results are based on the zero-shot test set. Primary experiments were conducted on four Nvidia A100-80GB GPUs. More implementation details are in Appendix~\ref{sec:appendixA}.

\subsection{Main Results}
Main results are shown in Table~\ref{table_results}. \\
\textbf{Advantage of Distillation} \ \ The inference results of student LM Llama2-7B show significant improvement after applying knowledge distillation. Although it still has a noticeable gap between the distilled Llama2-7B and teacher LLMs in mathematical reasoning after distillation, the fine-tuned Llama2-7B outperforms the weakest teacher LLM in commonsense and logical tasks. As more updated and powerful student LMs, Qwen2.5-1.5B and Llama3.1-8B show steady improvements after distillation. Notably, the multiple-teacher distillation results on Llama3.1-8B even surpass all teacher LLMs. Considering that we only use the failed cases set as shown in Table~\ref{table_exam}, it demonstrates that FAIR method effectively integrates LLMs to enhance the reasoning abilities of student models. 
\begin{table}[htb]
\centering
\resizebox{1\columnwidth}{!}{\begin{tabular}{lllll}
\toprule \multicolumn{5}{c}{\textbf{\#Wrong / \#Train}} \\ 
Model & GSM8K & SVAMP & StrategyQA & LogiQA\\ \midrule
Qwen2.5-1.5B & 1705/7473 & 136/700 & 671/1603 & 5450/7376\\
Llama2-7B & 6236/7473 & 387/700 & 825/1603 & 6159/7376\\
Llama3.1-8B & 1422/7473 & 132/700 & 554/1603 & 4433/7376\\ \bottomrule
\end{tabular}}
\caption{\label{table_exam}Exam results on original student models. The wrongly answered samples will be collected for generating the teacher responses and distillation training set.}
\vspace{-0.5cm}
\end{table} \\
\textbf{Comparison with Baselines} \ \ Compared to distillation methods on smaller models such as CodeT5, Qwen2-1.5B , and GPT-J, FAIR on Qwen2.5-1.5B consistently achieves superior performance on the available mathematical and commonsense tasks. Compared with other works based on Llama-series models, on the GSM8K benchmark, our performance on Llama2-7B (36.24\%) lags behind Llama-7B+NCE (41.93\%) and ReversalMath (52.10\%), likely because these models were exclusively fine-tuned on mathematical tasks, with GSM8K being a key and difficult benchmark in this domain. The other trained mathematical datasets may improve student LM's overall mathematical reasoning capability. In addition, we utilize only the failed cases set, which is significantly smaller compared to the training data in other studies. Nevertheless, our approach still yields better performance compared to ReversalMath on another easier and smaller mathematical benchmark, SVAMP (59.50\%>59.20\%). Additionally, our results on LogiQA (36.25\%) also exceed the ORCA2-7B (35.02\%). Finally, distillation results on Llama3.1-8B-Instruct surpass the same Llama3.1-8B-Instruct+ReDistill and the larger T5-XXL+CoT on mathematical and commonsense tasks. In conclusion, despite using less powerful teacher models, our method still outperforms related work that leverages state-of-the-art LLMs such as GPT-4 and DeepSeek-R1.

\section{Analysis}
\subsection{Analysis about Peer-Review Process}
To assess the importance of the peer-review process further, we compared the evaluation results with and without peer-review, as shown in Table~\ref{table_results}. When peer-review is absent, the average test accuracy across all benchmarks decreases by 7.84\%, 5.18\%, and 2.83\% for Llama2-7B, Qwen2.5-1.5B, and Llama3.1-8B, respectively. This reinforces that noisy answers generated by multiple teachers, which could potentially confuse the student model during instruction tuning, can be effectively filtered through peer review, ultimately enhancing the student model's performance.
In addition, for our backbone Llama2-7B, the experiments without peer-review even fall behind the best single teacher-GPT distillation outcomes on GSM8K (29.65\%<30.71\%). This pattern is particularly pronounced in commonsense and logical reasoning tasks. These findings align with our assumption that peer-review may have a smaller impact on mathematical reasoning tasks, where the rationale and final result are highly correlated, but significantly improves the quality of instruction data in commonsense and logical reasoning tasks. More results based on peer-review between only two teacher LLMs are displayed in Appendix~\ref{sec:appendix_2teachers}.

\subsection{Quality of Automated Peer-Review}
To further evaluate the reliability of our automated peer-review process, we conducted a manual analysis to assess whether the teachers' reasoning process genuinely supports their answers. This is important because an answer may sometimes be correct by chance despite flawed reasoning. First, we randomly selected 100 samples from {$D_{mistake}$} of the LogiQA dataset and collected the original ``correct'' responses, whenever a teacher model's predicted final answers matched the gold multiple-choice answers. We then manually examined these responses and removed those ``guessed'' correct answers with flawed rationales. Finally, we compared our gold-standard, human-annotated reasoning with those produced by the automated peer-review process. Table~\ref{table_human} revealed that the peer-review process achieved an average accuracy of \textbf{90.35\%} when compared to human annotations, demonstrating its high reliability.

\begin{table}[h]
\centering
\small
\resizebox{1\columnwidth}{!}{\begin{tabular}{llll}
\toprule 
Model & Original & PR & Human\\ \midrule
GPT-3.5-Turbo & 40 & 38 & 34\\
Gemini-1.0-Pro & 38 & 34 & 30\\
Mixtral-8x7B-Instruct  & 35 & 30 & 28\\ \bottomrule
\end{tabular}}
\caption{\label{table_human}Comparison of the number of responses verified by original model predictions, peer-review (PR), and human annotations for random 100 LogiQA samples.}
\vspace{-0.5cm}
\end{table}

\subsection{Abalation of Learning from Mistakes}
\begin{figure}[htb]
    \centering
     \includegraphics[width=0.45\textwidth,height=0.35\textwidth]{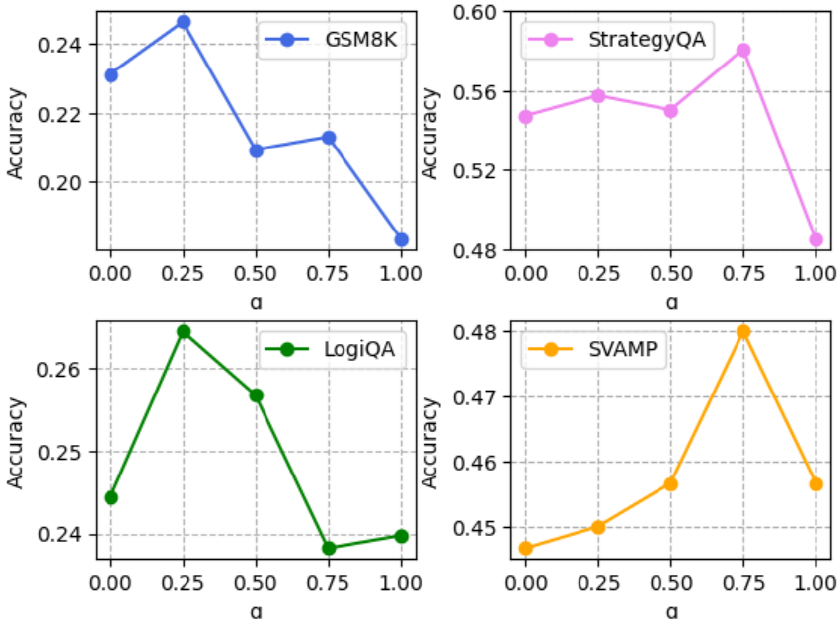}
     \vspace{-0.3cm}
     \caption{The effect of the tuning performance on Llama2-7B-chat. {$\alpha$}=0 indicates the absence of learning from mistakes.}
    \label{fig_alpha}
\vspace{-0.5cm}
\end{figure}

As a key component of our FAIR method, we initially set the proportion of learning from mistakes to 0.5 in previous experiments for simplicity. To explore the influence of balancing learning from rationales and learning from mistakes, we adjust the value of {$\alpha$} in Equation~\ref{eq:7}. Specifically, {$\alpha$} was varied from [0, 0.25, 0.5, 0.75, 1], and experiments were conducted on all benchmarks for 5 epochs on Llama2-7B-chat, while keeping other parameters constant. Figure~\ref{fig_alpha} visualizes how learning from mistakes affects instruction-tuning. Our findings support the hypothesis that learning from mistakes positively impacts instruction tuning. However, the relationship is not uniformly positive across all {$\alpha$} values on the four benchmarks.

For GSM8K and LogiQA, the benefits of learning from mistakes increase when {$\alpha<0.25$}, but start to decrease when {$\alpha$} exceeds 0.25. Conversely, for StrategyQA and SVAMP, the advantages of learning from mistakes consistently grow and reach their peak when {$\alpha=0.75$}. These results suggest that placing too much emphasis on learning from mistakes (i.e., a higher {$\alpha$} value) can lead to instability. Consequently, it is important to evaluate and optimize {$\alpha$} value for different tasks to effectively balance the learning of ``what'' (correct answers) and ``why'' (own mistakes) during training.

\subsection{Effectiveness of Multiple Teachers}
As shown in Table~\ref{table_results}, our multiple-teacher distillation with peer-review method on Llama2-7B improves the average accuracy by 5.48\% across four benchmarks compared to the single teacher distillation method with the highest accuracy. Although the performance gains on Qwen2.5-1.5B and Llama3.1-8B are slightly reduced, this is likely due to the strong baseline capabilities of the original student models, which are already competitive against teacher LLMs, and the limited size of the generated training set.

To ensure that all teacher LLMs contribute meaningfully to the final performance and prevent free-riding, Table~\ref{table_count} reports the number of responses utilized in the final multiple-teacher training tasks. They are generated by different LLMs and verified through the peer-review process. This comparison correlates with the distinct capabilities of each teacher model and underscores their collective contribution to enhancing the student model's performance after fine-tuning. Detailed comparisons of the student LM's output before and after distillation are provided in Appendix~\ref{sec:appendix_output}.

\begin{table}[h]
\centering
\resizebox{1\columnwidth}{!}{\begin{tabular}{llll}
\toprule
\textbf{Dataset} & \textbf{Qwen2.5-1.5B} & \textbf{Llama2-7B} & \textbf{Llama3.1-8B} \\ \midrule
\textbf{GSM8K} & 595:486:589 & 2110:1801:2256 & 472:420:506 \\
\textbf{SVAMP} & 34:28:63 & 161:87:117 & 31:27:64 \\
\textbf{StrategyQA} & 333:143:123 & 396:192:160 & 293:106:78 \\
\textbf{LogiQA} & 1731:1884:1545 & 1706:2085:2049 & 1478:1403:1286 \\
 \bottomrule
\end{tabular}}
\caption{\label{table_count}The number of responses from various teacher LLMs used in the final multiple-teacher distillation process. The values represent the number of data points from Mixtral/Gemini/GPT respectively. This demonstrates that all teacher LLMs contribute significantly.}
\vspace{-0.5cm}
\end{table}

\subsection{Assessment of Computational Overhead}
To address concerns about the additional computational overhead introduced by FAIR, we evaluate the resources consumed during our experiments. Table~\ref{table_cost} provides a comparison of the average number of tokens consumed for each sample with and without the peer-review. 
The selected teacher models are all entry-level LLMs that do not require subscriptions or high costs, ensuring accessibility for researchers with limited resources. 
Given the substantial improvement in the student model’s performance and the fact that distillation is a one-time investment, the additional cost is highly justified. Moreover, the distilled model can even outperform certain teacher LLMs on specific benchmarks while maintaining significantly lower inference costs.

\begin{table}[h]
\centering
\small
\resizebox{1\columnwidth}{!}{\begin{tabular}{llll}
\toprule
LLMs & Standard & PR & $\Delta$Cost $\uparrow$\\ \midrule
GPT-3.5-Turbo & 100.32 & 200.18 & \$0.00005\\
Gemini-1.0-Pro & 60.94 & 220.82 & \$0.00008\\
Mixtral-8x7B-Instruct & 131.12 & 214.45 & \$0.00002\\ \bottomrule
\end{tabular}}
\caption{\label{table_cost}The average number of tokens consumed for each sample with and without the peer-review (PR).}
\vspace{-0.7cm}
\end{table}

\section{Conclusion}
In this work, we introduce the Fault-Aware Distillation via Peer-Review (FAIR) approach. We implement a simulated peer-review process between multiple teacher LLMs to gather reliable outputs, which refines the quality of instruction tuning dataset. Additionally, we develop an integrated instruction tuning method that allows the student LM to learn from both the correct rationale and mistakes feedback. Comprehensive results on diverse reasoning tasks validate our efficient method for unlocking the reasoning potential of smaller open-source LMs through distillation, even with black-box LLMs and without dataset-provided rationales. We hope that our findings will encourage further investigations into reasoning distillation.

\section*{Limitations}
Although our method demonstrates effectiveness in the reasoning ability distillation from teacher models to the student model, this technique has several limitations. First, our experiments primarily rely on GPT-3.5-Turbo, Gemini-1.0-Pro, and Mixtral-8x7B-Instruct-v0.1 as teacher LLMs due to considerations of availability and cost. The results in Table~\ref{table_results} suggest that as student models improve, the bottleneck in performance may shift to the capabilities of the teacher LLMs, highlighting the need for more advanced teacher models to further enhance student performance. Future research could benefit from using more powerful models like DeepSeek-R1, OpenAI-o3, and Claude-3 Opus. 
Secondly, future work could include more challenging benchmarks across different reasoning fields, such as FrontierMath~\cite{glazer2024frontiermath} and Humanity's Last Exam~\cite{phan2025humanity}.
Thirdly, due to time and cost constraints, our method does not collect the student LM's incorrect rationales and updates the instruction dataset after each epoch. The potential benefits of continuously incorporating fresh data throughout online training remain unexplored. Moreover, further research can regard teacher LLMs as agents, incorporating more sophisticated pipelines such as negotiation and decision-making during the peer-review process to enhance reliability.
Lastly, we employ the default cross-entropy loss function for instruction tuning. It would be worthwhile to explore more sophisticated methods, such as the Group Relative Policy Optimization (GRPO) Reinforcement Learning method used in DeepSeek-R1, and additional techniques.

\section*{Ethics Statement}
The study offers a novel structure for knowledge distillation of the reasoning ability from LLMs to smaller LM, which could contribute to increased transparency and availability in AI systems. It underscores the fact that proprietary LLMs dominate reasoning tasks and weaken smaller open-source LMs. However, parts of the annotated data in this paper are collected from close-source GPT provided by OpenAI, and Gemini supplied by Google. The explainability and transparency of close-source models may raise risks for annotated data and decrease the trustworthiness.

\section*{Acknowledgements}
We thank the reviewers for their valuable feedback. We also thank Xiang(Lorraine) Li, Joey Hou, Bhiman Kumar Baghel, Alejandro Ciuba, and Arun Balajiee for useful comments on an earlier draft of the paper. The infrastructure for all experiments is supported by The University of Pittsburgh Center for Research Computing (Pitt CRC) and Pittsburgh Supercomputing Center (PSC) Bridges2 HPC Resource.

\bibliography{anthology,custom}

\begin{thebibliography}{58}
\expandafter\ifx\csname natexlab\endcsname\relax\def\natexlab#1{#1}\fi

\bibitem[{Achiam et~al.(2023)Achiam, Adler, Agarwal, Ahmad, Akkaya, Aleman, Almeida, Altenschmidt, Altman, Anadkat et~al.}]{achiam2023gpt}
Josh Achiam, Steven Adler, Sandhini Agarwal, Lama Ahmad, Ilge Akkaya, Florencia~Leoni Aleman, Diogo Almeida, Janko Altenschmidt, Sam Altman, Shyamal Anadkat, et~al. 2023.
\newblock Gpt-4 technical report.
\newblock \emph{arXiv preprint arXiv:2303.08774}.

\bibitem[{Adarsh et~al.(2025)Adarsh, Shridhar, Gulcehre, Monath, and Sachan}]{adarsh2025siked}
Shivam Adarsh, Kumar Shridhar, Caglar Gulcehre, Nicholas Monath, and Mrinmaya Sachan. 2025.
\newblock \href {https://openreview.net/forum?id=ozTREVBARB} {{SIK}ed: Self-guided iterative knowledge distillation for mathematical reasoning}.

\bibitem[{Chen et~al.(2023)Chen, Wu, Quan, Wang, Yan, and Zhang}]{chen2023mcc}
Hongzhan Chen, Siyue Wu, Xiaojun Quan, Rui Wang, Ming Yan, and Ji~Zhang. 2023.
\newblock Mcc-kd: Multi-cot consistent knowledge distillation.
\newblock \emph{arXiv preprint arXiv:2310.14747}.

\bibitem[{Chenglin et~al.(2023)Chenglin, Qianglong, Caiyu, and Yin}]{chenglin2023mixed}
Li~Chenglin, Chen Qianglong, Wang Caiyu, and Zhang Yin. 2023.
\newblock Mixed distillation helps smaller language model better reasoning.
\newblock \emph{arXiv preprint arXiv:2312.10730}.

\bibitem[{Chu et~al.(2023)Chu, Chen, Chen, Yu, He, Wang, Peng, Liu, Qin, and Liu}]{chu2023survey}
Zheng Chu, Jingchang Chen, Qianglong Chen, Weijiang Yu, Tao He, Haotian Wang, Weihua Peng, Ming Liu, Bing Qin, and Ting Liu. 2023.
\newblock A survey of chain of thought reasoning: Advances, frontiers and future.
\newblock \emph{arXiv preprint arXiv:2309.15402}.

\bibitem[{Chu et~al.(2024)Chu, Ai, Tu, Li, and Liu}]{chu2024pre}
Zhumin Chu, Qingyao Ai, Yiteng Tu, Haitao Li, and Yiqun Liu. 2024.
\newblock Pre: A peer review based large language model evaluator.
\newblock \emph{arXiv preprint arXiv:2401.15641}.

\bibitem[{Chung et~al.(2024)Chung, Hou, Longpre, Zoph, Tay, Fedus, Li, Wang, Dehghani, Brahma et~al.}]{chung2024scaling}
Hyung~Won Chung, Le~Hou, Shayne Longpre, Barret Zoph, Yi~Tay, William Fedus, Yunxuan Li, Xuezhi Wang, Mostafa Dehghani, Siddhartha Brahma, et~al. 2024.
\newblock Scaling instruction-finetuned language models.
\newblock \emph{Journal of Machine Learning Research}, 25(70):1--53.

\bibitem[{Cobbe et~al.(2021)Cobbe, Kosaraju, Bavarian, Chen, Jun, Kaiser, Plappert, Tworek, Hilton, Nakano et~al.}]{cobbe2021training}
Karl Cobbe, Vineet Kosaraju, Mohammad Bavarian, Mark Chen, Heewoo Jun, Lukasz Kaiser, Matthias Plappert, Jerry Tworek, Jacob Hilton, Reiichiro Nakano, et~al. 2021.
\newblock Training verifiers to solve math word problems.
\newblock \emph{arXiv preprint arXiv:2110.14168}.

\bibitem[{Dao et~al.(2022)Dao, Fu, Ermon, Rudra, and R{\'e}}]{dao2022flashattention}
Tri Dao, Dan Fu, Stefano Ermon, Atri Rudra, and Christopher R{\'e}. 2022.
\newblock Flashattention: Fast and memory-efficient exact attention with io-awareness.
\newblock \emph{Advances in Neural Information Processing Systems}, 35:16344--16359.

\bibitem[{Dubey et~al.(2024)Dubey, Jauhri, Pandey, Kadian, Al-Dahle, Letman, Mathur, Schelten, Yang, Fan et~al.}]{dubey2024llama}
Abhimanyu Dubey, Abhinav Jauhri, Abhinav Pandey, Abhishek Kadian, Ahmad Al-Dahle, Aiesha Letman, Akhil Mathur, Alan Schelten, Amy Yang, Angela Fan, et~al. 2024.
\newblock The llama 3 herd of models.
\newblock \emph{arXiv preprint arXiv:2407.21783}.

\bibitem[{Fan and Tao(2024)}]{fan2024towards}
Xiaojing Fan and Chunliang Tao. 2024.
\newblock Towards resilient and efficient llms: A comparative study of efficiency, performance, and adversarial robustness.
\newblock \emph{arXiv preprint arXiv:2408.04585}.

\bibitem[{Fu et~al.(2023)Fu, Peng, Ou, Sabharwal, and Khot}]{fu2023specializing}
Yao Fu, Hao Peng, Litu Ou, Ashish Sabharwal, and Tushar Khot. 2023.
\newblock Specializing smaller language models towards multi-step reasoning.
\newblock In \emph{International Conference on Machine Learning}, pages 10421--10430. PMLR.

\bibitem[{Geva et~al.(2021)Geva, Khashabi, Segal, Khot, Roth, and Berant}]{geva2021didaristotleuselaptop}
Mor Geva, Daniel Khashabi, Elad Segal, Tushar Khot, Dan Roth, and Jonathan Berant. 2021.
\newblock \href {http://arxiv.org/abs/2101.02235} {Did aristotle use a laptop? a question answering benchmark with implicit reasoning strategies}.

\bibitem[{Glazer et~al.(2024)Glazer, Erdil, Besiroglu, Chicharro, Chen, Gunning, Olsson, Denain, Ho, Santos et~al.}]{glazer2024frontiermath}
Elliot Glazer, Ege Erdil, Tamay Besiroglu, Diego Chicharro, Evan Chen, Alex Gunning, Caroline~Falkman Olsson, Jean-Stanislas Denain, Anson Ho, Emily de~Oliveira Santos, et~al. 2024.
\newblock Frontiermath: A benchmark for evaluating advanced mathematical reasoning in ai.
\newblock \emph{arXiv preprint arXiv:2411.04872}.

\bibitem[{Guo et~al.(2025)Guo, Yang, Zhang, Song, Zhang, Xu, Zhu, Ma, Wang, Bi et~al.}]{guo2025deepseek}
Daya Guo, Dejian Yang, Haowei Zhang, Junxiao Song, Ruoyu Zhang, Runxin Xu, Qihao Zhu, Shirong Ma, Peiyi Wang, Xiao Bi, et~al. 2025.
\newblock Deepseek-r1: Incentivizing reasoning capability in llms via reinforcement learning.
\newblock \emph{arXiv preprint arXiv:2501.12948}.

\bibitem[{Guo et~al.(2024)Guo, You, Li, Bowen, and Zhang}]{guo2024exploring}
Pei Guo, Wangjie You, Juntao Li, Yan Bowen, and Min Zhang. 2024.
\newblock Exploring reversal mathematical reasoning ability for large language models.
\newblock In \emph{Findings of the Association for Computational Linguistics ACL 2024}, pages 13671--13685.

\bibitem[{Gurrapu et~al.(2023)Gurrapu, Kulkarni, Huang, Lourentzou, and Batarseh}]{gurrapu2023rationalization}
Sai Gurrapu, Ajay Kulkarni, Lifu Huang, Ismini Lourentzou, and Feras~A Batarseh. 2023.
\newblock Rationalization for explainable nlp: a survey.
\newblock \emph{Frontiers in Artificial Intelligence}, 6:1225093.

\bibitem[{Hicham~Badri(2025)}]{badri2025r1}
Appu~Shaji Hicham~Badri. 2025.
\newblock \href {https://mobiusml.github.io/r1_redistill_blogpost/} {Re-distilling smaller deepseek r1 models for better performance}.

\bibitem[{Ho et~al.(2023)Ho, Schmid, and Yun}]{ho2023large}
Namgyu Ho, Laura Schmid, and Se-Young Yun. 2023.
\newblock Large language models are reasoning teachers.
\newblock In \emph{Proceedings of the 61st Annual Meeting of the Association for Computational Linguistics (Volume 1: Long Papers)}, pages 14852--14882.

\bibitem[{Huang et~al.(2022)Huang, Gu, Hou, Wu, Wang, Yu, and Han}]{huang2022large}
Jiaxin Huang, Shixiang~Shane Gu, Le~Hou, Yuexin Wu, Xuezhi Wang, Hongkun Yu, and Jiawei Han. 2022.
\newblock Large language models can self-improve.
\newblock \emph{arXiv preprint arXiv:2210.11610}.

\bibitem[{Imani et~al.(2023)Imani, Du, and Shrivastava}]{imani2023mathprompter}
Shima Imani, Liang Du, and Harsh Shrivastava. 2023.
\newblock Mathprompter: Mathematical reasoning using large language models.
\newblock \emph{arXiv preprint arXiv:2303.05398}.

\bibitem[{Jiang et~al.(2024)Jiang, Sablayrolles, Roux, Mensch, Savary, Bamford, Chaplot, Casas, Hanna, Bressand et~al.}]{jiang2024mixtral}
Albert~Q Jiang, Alexandre Sablayrolles, Antoine Roux, Arthur Mensch, Blanche Savary, Chris Bamford, Devendra~Singh Chaplot, Diego de~las Casas, Emma~Bou Hanna, Florian Bressand, et~al. 2024.
\newblock Mixtral of experts.
\newblock \emph{arXiv preprint arXiv:2401.04088}.

\bibitem[{Jiang et~al.(2023)Jiang, Chan, Chen, and Wang}]{jiang2023lion}
Yuxin Jiang, Chunkit Chan, Mingyang Chen, and Wei Wang. 2023.
\newblock Lion: Adversarial distillation of proprietary large language models.
\newblock \emph{arXiv preprint arXiv:2305.12870}.

\bibitem[{Konold et~al.(2004)Konold, Miller, and Konold}]{konold2004using}
Kathryn~E Konold, Susan~P Miller, and Kyle~B Konold. 2004.
\newblock Using teacher feedback to enhance student learning.
\newblock \emph{Teaching Exceptional Children}, 36(6):64--69.

\bibitem[{Lee et~al.(2024)Lee, Kim, and Lee}]{lee2024mentor}
Hojae Lee, Junho Kim, and SangKeun Lee. 2024.
\newblock Mentor-kd: Making small language models better multi-step reasoners.
\newblock \emph{arXiv preprint arXiv:2410.09037}.

\bibitem[{Lewkowycz et~al.(2022)Lewkowycz, Andreassen, Dohan, Dyer, Michalewski, Ramasesh, Slone, Anil, Schlag, Gutman-Solo et~al.}]{lewkowycz2022solving}
Aitor Lewkowycz, Anders Andreassen, David Dohan, Ethan Dyer, Henryk Michalewski, Vinay Ramasesh, Ambrose Slone, Cem Anil, Imanol Schlag, Theo Gutman-Solo, et~al. 2022.
\newblock Solving quantitative reasoning problems with language models.
\newblock \emph{Advances in Neural Information Processing Systems}, 35:3843--3857.

\bibitem[{Li et~al.(2024{\natexlab{a}})Li, Chen, Wang, Zhao, Liang, Hou, Liu, and Zhou}]{li2024mosaic}
Ming Li, Pei Chen, Chenguang Wang, Hongyu Zhao, Yijun Liang, Yupeng Hou, Fuxiao Liu, and Tianyi Zhou. 2024{\natexlab{a}}.
\newblock Mosaic-it: Free compositional data augmentation improves instruction tuning.
\newblock \emph{arXiv preprint arXiv:2405.13326}.

\bibitem[{Li et~al.(2024{\natexlab{b}})Li, Yuan, Feng, Pan, Sun, Wang, Wang, and Li}]{li2024turning}
Yiwei Li, Peiwen Yuan, Shaoxiong Feng, Boyuan Pan, Bin Sun, Xinglin Wang, Heda Wang, and Kan Li. 2024{\natexlab{b}}.
\newblock Turning dust into gold: Distilling complex reasoning capabilities from llms by leveraging negative data.
\newblock In \emph{Proceedings of the AAAI Conference on Artificial Intelligence}, volume~38, pages 18591--18599.

\bibitem[{Liu et~al.(2023)Liu, Ning, Teng, Liu, Zhou, and Zhang}]{liu2023evaluating}
Hanmeng Liu, Ruoxi Ning, Zhiyang Teng, Jian Liu, Qiji Zhou, and Yue Zhang. 2023.
\newblock Evaluating the logical reasoning ability of chatgpt and gpt-4.
\newblock \emph{arXiv preprint arXiv:2304.03439}.

\bibitem[{Liu et~al.(2020)Liu, Cui, Liu, Huang, Wang, and Zhang}]{liu2020logiqachallengedatasetmachine}
Jian Liu, Leyang Cui, Hanmeng Liu, Dandan Huang, Yile Wang, and Yue Zhang. 2020.
\newblock \href {http://arxiv.org/abs/2007.08124} {Logiqa: A challenge dataset for machine reading comprehension with logical reasoning}.

\bibitem[{Magister et~al.(2022)Magister, Mallinson, Adamek, Malmi, and Severyn}]{magister2022teaching}
Lucie~Charlotte Magister, Jonathan Mallinson, Jakub Adamek, Eric Malmi, and Aliaksei Severyn. 2022.
\newblock Teaching small language models to reason.
\newblock \emph{arXiv preprint arXiv:2212.08410}.

\bibitem[{Mitra et~al.(2023)Mitra, Del~Corro, Mahajan, Codas, Simoes, Agarwal, Chen, Razdaibiedina, Jones, Aggarwal et~al.}]{mitra2023orca}
Arindam Mitra, Luciano Del~Corro, Shweti Mahajan, Andres Codas, Clarisse Simoes, Sahaj Agarwal, Xuxi Chen, Anastasia Razdaibiedina, Erik Jones, Kriti Aggarwal, et~al. 2023.
\newblock Orca 2: Teaching small language models how to reason.
\newblock \emph{arXiv preprint arXiv:2311.11045}.

\bibitem[{Muennighoff et~al.(2025)Muennighoff, Yang, Shi, Li, Fei-Fei, Hajishirzi, Zettlemoyer, Liang, Cand{\`e}s, and Hashimoto}]{muennighoff2025s1}
Niklas Muennighoff, Zitong Yang, Weijia Shi, Xiang~Lisa Li, Li~Fei-Fei, Hannaneh Hajishirzi, Luke Zettlemoyer, Percy Liang, Emmanuel Cand{\`e}s, and Tatsunori Hashimoto. 2025.
\newblock s1: Simple test-time scaling.
\newblock \emph{arXiv preprint arXiv:2501.19393}.

\bibitem[{Nan et~al.(2023)Nan, Zhang, Zou, Zhao, Zhou, and Cohan}]{nan2023evaluating}
Linyong Nan, Ellen Zhang, Weijin Zou, Yilun Zhao, Wenfei Zhou, and Arman Cohan. 2023.
\newblock On evaluating the integration of reasoning and action in llm agents with database question answering.
\newblock \emph{arXiv preprint arXiv:2311.09721}.

\bibitem[{Ni et~al.(2024)Ni, Meng, Chen, Zhao, Chen, Li, Zhang, Yin, Wang, and Chan}]{ni2024earnings}
Haowei Ni, Shuchen Meng, Xupeng Chen, Ziqing Zhao, Andi Chen, Panfeng Li, Shiyao Zhang, Qifu Yin, Yuanqing Wang, and Yuxi Chan. 2024.
\newblock Harnessing earnings reports for stock predictions: A qlora-enhanced llm approach.
\newblock \emph{arXiv preprint arXiv:2408.06634}.

\bibitem[{Ning et~al.(2024)Ning, Yang, Liu, Yao, Liu, Wang, Pang, and Yuan}]{ning2024peer}
Kun-Peng Ning, Shuo Yang, Yu-Yang Liu, Jia-Yu Yao, Zhen-Hui Liu, Yu~Wang, Ming Pang, and Li~Yuan. 2024.
\newblock Peer-review-in-llms: Automatic evaluation method for llms in open-environment.
\newblock \emph{arXiv preprint arXiv:2402.01830}.

\bibitem[{Patel et~al.(2021)Patel, Bhattamishra, and Goyal}]{patel2021nlp}
Arkil Patel, Satwik Bhattamishra, and Navin Goyal. 2021.
\newblock Are nlp models really able to solve simple math word problems?
\newblock \emph{arXiv preprint arXiv:2103.07191}.

\bibitem[{Phan et~al.(2025)Phan, Gatti, Han, Li, Hu, Zhang, Shi, Choi, Agrawal, Chopra et~al.}]{phan2025humanity}
Long Phan, Alice Gatti, Ziwen Han, Nathaniel Li, Josephina Hu, Hugh Zhang, Sean Shi, Michael Choi, Anish Agrawal, Arnav Chopra, et~al. 2025.
\newblock Humanity's last exam.
\newblock \emph{arXiv preprint arXiv:2501.14249}.

\bibitem[{Shridhar et~al.(2022)Shridhar, Stolfo, and Sachan}]{shridhar2022distilling}
Kumar Shridhar, Alessandro Stolfo, and Mrinmaya Sachan. 2022.
\newblock Distilling reasoning capabilities into smaller language models.
\newblock \emph{arXiv preprint arXiv:2212.00193}.

\bibitem[{Sun et~al.(2023)Sun, Yin, Li, Wu, Qiu, and Kong}]{sun2023corex}
Qiushi Sun, Zhangyue Yin, Xiang Li, Zhiyong Wu, Xipeng Qiu, and Lingpeng Kong. 2023.
\newblock Corex: Pushing the boundaries of complex reasoning through multi-model collaboration.
\newblock \emph{arXiv preprint arXiv:2310.00280}.

\bibitem[{Team et~al.(2023)Team, Anil, Borgeaud, Wu, Alayrac, Yu, Soricut, Schalkwyk, Dai, Hauth et~al.}]{team2023gemini}
Gemini Team, Rohan Anil, Sebastian Borgeaud, Yonghui Wu, Jean-Baptiste Alayrac, Jiahui Yu, Radu Soricut, Johan Schalkwyk, Andrew~M Dai, Anja Hauth, et~al. 2023.
\newblock Gemini: a family of highly capable multimodal models.
\newblock \emph{arXiv preprint arXiv:2312.11805}.

\bibitem[{Tian et~al.(2024)Tian, Han, Chen, Wang, and Chawla}]{tian2024tinyllm}
Yijun Tian, Yikun Han, Xiusi Chen, Wei Wang, and Nitesh~V Chawla. 2024.
\newblock Tinyllm: Learning a small student from multiple large language models.
\newblock \emph{arXiv preprint arXiv:2402.04616}.

\bibitem[{Touvron et~al.(2023)Touvron, Martin, Stone, Albert, Almahairi, Babaei, Bashlykov, Batra, Bhargava, Bhosale et~al.}]{touvron2023llama}
Hugo Touvron, Louis Martin, Kevin Stone, Peter Albert, Amjad Almahairi, Yasmine Babaei, Nikolay Bashlykov, Soumya Batra, Prajjwal Bhargava, Shruti Bhosale, et~al. 2023.
\newblock Llama 2: Open foundation and fine-tuned chat models.
\newblock \emph{arXiv preprint arXiv:2307.09288}.

\bibitem[{Wang et~al.(2022)Wang, Kordi, Mishra, Liu, Smith, Khashabi, and Hajishirzi}]{wang2022self}
Yizhong Wang, Yeganeh Kordi, Swaroop Mishra, Alisa Liu, Noah~A Smith, Daniel Khashabi, and Hannaneh Hajishirzi. 2022.
\newblock Self-instruct: Aligning language models with self-generated instructions.
\newblock \emph{arXiv preprint arXiv:2212.10560}.

\bibitem[{Wang et~al.(2023)Wang, Huang, Liu, Wang, Song, Zhang, Huang, Wei, Deng, Sun et~al.}]{wang2023democratizing}
Zhaoyang Wang, Shaohan Huang, Yuxuan Liu, Jiahai Wang, Minghui Song, Zihan Zhang, Haizhen Huang, Furu Wei, Weiwei Deng, Feng Sun, et~al. 2023.
\newblock Democratizing reasoning ability: Tailored learning from large language model.
\newblock \emph{arXiv preprint arXiv:2310.13332}.

\bibitem[{Wei et~al.(2021)Wei, Bosma, Zhao, Guu, Yu, Lester, Du, Dai, and Le}]{wei2021finetuned}
Jason Wei, Maarten Bosma, Vincent~Y Zhao, Kelvin Guu, Adams~Wei Yu, Brian Lester, Nan Du, Andrew~M Dai, and Quoc~V Le. 2021.
\newblock Finetuned language models are zero-shot learners.
\newblock \emph{arXiv preprint arXiv:2109.01652}.

\bibitem[{Wei et~al.(2022{\natexlab{a}})Wei, Tay, Bommasani, Raffel, Zoph, Borgeaud, Yogatama, Bosma, Zhou, Metzler et~al.}]{wei2022emergent}
Jason Wei, Yi~Tay, Rishi Bommasani, Colin Raffel, Barret Zoph, Sebastian Borgeaud, Dani Yogatama, Maarten Bosma, Denny Zhou, Donald Metzler, et~al. 2022{\natexlab{a}}.
\newblock Emergent abilities of large language models.
\newblock \emph{arXiv preprint arXiv:2206.07682}.

\bibitem[{Wei et~al.(2022{\natexlab{b}})Wei, Wang, Schuurmans, Bosma, Xia, Chi, Le, Zhou et~al.}]{wei2022chain}
Jason Wei, Xuezhi Wang, Dale Schuurmans, Maarten Bosma, Fei Xia, Ed~Chi, Quoc~V Le, Denny Zhou, et~al. 2022{\natexlab{b}}.
\newblock Chain-of-thought prompting elicits reasoning in large language models.
\newblock \emph{Advances in neural information processing systems}, 35:24824--24837.

\bibitem[{Xu et~al.(2023{\natexlab{a}})Xu, Guo, Duan, and McAuley}]{xu2023baize}
Canwen Xu, Daya Guo, Nan Duan, and Julian McAuley. 2023{\natexlab{a}}.
\newblock Baize: An open-source chat model with parameter-efficient tuning on self-chat data.
\newblock \emph{arXiv preprint arXiv:2304.01196}.

\bibitem[{Xu et~al.(2023{\natexlab{b}})Xu, Lin, Han, Zhao, Liu, and Cambria}]{xu2023large}
Fangzhi Xu, Qika Lin, Jiawei Han, Tianzhe Zhao, Jun Liu, and Erik Cambria. 2023{\natexlab{b}}.
\newblock Are large language models really good logical reasoners? a comprehensive evaluation from deductive, inductive and abductive views.
\newblock \emph{arXiv preprint arXiv:2306.09841}.

\bibitem[{Xu et~al.(2024{\natexlab{a}})Xu, Ye, Li, and Chen}]{xu2024can}
Han Xu, Jingyang Ye, Yutong Li, and Haipeng Chen. 2024{\natexlab{a}}.
\newblock \href {https://openreview.net/forum?id=42b9hJrIpX} {Can speculative sampling accelerate react without compromising reasoning quality?}
\newblock In \emph{The Second Tiny Papers Track at ICLR 2024}.

\bibitem[{Xu et~al.(2024{\natexlab{b}})Xu, Li, Tao, Shen, Cheng, Li, Xu, Tao, and Zhou}]{xu2024survey}
Xiaohan Xu, Ming Li, Chongyang Tao, Tao Shen, Reynold Cheng, Jinyang Li, Can Xu, Dacheng Tao, and Tianyi Zhou. 2024{\natexlab{b}}.
\newblock A survey on knowledge distillation of large language models.
\newblock \emph{arXiv preprint arXiv:2402.13116}.

\bibitem[{Yang et~al.(2024)Yang, Yang, Hui, Zheng, Yu, Zhou, Li, Li, Liu, Huang, Dong, Wei, Lin, Tang, Wang, Yang, Tu, Zhang, Ma, Xu, Zhou, Bai, He, Lin, Dang, Lu, Chen, Yang, Li, Xue, Ni, Zhang, Wang, Peng, Men, Gao, Lin, Wang, Bai, Tan, Zhu, Li, Liu, Ge, Deng, Zhou, Ren, Zhang, Wei, Ren, Fan, Yao, Zhang, Wan, Chu, Liu, Cui, Zhang, and Fan}]{qwen2}
An~Yang, Baosong Yang, Binyuan Hui, Bo~Zheng, Bowen Yu, Chang Zhou, Chengpeng Li, Chengyuan Li, Dayiheng Liu, Fei Huang, Guanting Dong, Haoran Wei, Huan Lin, Jialong Tang, Jialin Wang, Jian Yang, Jianhong Tu, Jianwei Zhang, Jianxin Ma, Jin Xu, Jingren Zhou, Jinze Bai, Jinzheng He, Junyang Lin, Kai Dang, Keming Lu, Keqin Chen, Kexin Yang, Mei Li, Mingfeng Xue, Na~Ni, Pei Zhang, Peng Wang, Ru~Peng, Rui Men, Ruize Gao, Runji Lin, Shijie Wang, Shuai Bai, Sinan Tan, Tianhang Zhu, Tianhao Li, Tianyu Liu, Wenbin Ge, Xiaodong Deng, Xiaohuan Zhou, Xingzhang Ren, Xinyu Zhang, Xipin Wei, Xuancheng Ren, Yang Fan, Yang Yao, Yichang Zhang, Yu~Wan, Yunfei Chu, Yuqiong Liu, Zeyu Cui, Zhenru Zhang, and Zhihao Fan. 2024.
\newblock Qwen2 technical report.
\newblock \emph{arXiv preprint arXiv:2407.10671}.

\bibitem[{Zelikman et~al.(2022)Zelikman, Wu, Mu, and Goodman}]{zelikman2022star}
Eric Zelikman, Yuhuai Wu, Jesse Mu, and Noah Goodman. 2022.
\newblock \href {https://openreview.net/forum?id=_3ELRdg2sgI} {{ST}ar: Bootstrapping reasoning with reasoning}.
\newblock In \emph{Advances in Neural Information Processing Systems}.

\bibitem[{Zeng et~al.(2022)Zeng, Liu, Du, Wang, Lai, Ding, Yang, Xu, Zheng, Xia et~al.}]{zeng2022glm}
Aohan Zeng, Xiao Liu, Zhengxiao Du, Zihan Wang, Hanyu Lai, Ming Ding, Zhuoyi Yang, Yifan Xu, Wendi Zheng, Xiao Xia, et~al. 2022.
\newblock Glm-130b: An open bilingual pre-trained model.
\newblock \emph{arXiv preprint arXiv:2210.02414}.

\bibitem[{Zhao et~al.(2024)Zhao, Lee, and Hsu}]{zhao2024large}
Zirui Zhao, Wee~Sun Lee, and David Hsu. 2024.
\newblock Large language models as commonsense knowledge for large-scale task planning.
\newblock \emph{Advances in Neural Information Processing Systems}, 36.

\bibitem[{Zhou and Ai(2024)}]{zhou2024teaching}
Yuhang Zhou and Wei Ai. 2024.
\newblock Teaching-assistant-in-the-loop: Improving knowledge distillation from imperfect teacher models in low-budget scenarios.
\newblock \emph{arXiv preprint arXiv:2406.05322}.

\bibitem[{Zhu et~al.(2024)Zhu, Qi, Zhang, Long, Lin, and Zhou}]{zhu2024pad}
Xuekai Zhu, Biqing Qi, Kaiyan Zhang, Xinwei Long, Zhouhan Lin, and Bowen Zhou. 2024.
\newblock Pad: Program-aided distillation can teach small models reasoning better than chain-of-thought fine-tuning.
\newblock In \emph{Proceedings of the 2024 Conference of the North American Chapter of the Association for Computational Linguistics: Human Language Technologies (Volume 1: Long Papers)}, pages 2571--2597.

\end{thebibliography}
\bibliographystyle{acl_natbib}
\newpage

\appendix

\section{Experimental Setup Details}
\label{sec:appendixA}
\subsection{Datasets Statistics}
\label{sec:appendixA.1}
We download datasets GSM8K, SVAMP, StrategyQA, and LogiQA from Huggingface. All datasets are split according to the official original split ratio. Table~\ref{table_dataset} shows the dataset statistics.
\begin{table}[ht]
\centering
{\begin{tabular}{llcc}
\toprule Dataset & Type & \#Train & \#Test\\ \midrule
GSM8K & Mathematical & 7473  & 1319\\
SVAMP & Mathematical & 700  & 300\\
StrategyQA & Commonsense & 1603  & 687\\
LogiQA & Logical & 7376  & 651\\ \bottomrule
\end{tabular}}
\caption{\label{table_dataset} Dataset statistics. }
\end{table}

\subsection{Teacher LLMs Parameters}
Table~\ref{table_teacher_parse} shows the unified parameters setting for GPT-3.5-Turbo, Gemini-1.0-Pro, and Mixtral-8x7B-Instruct-v0.1 LLMs to generate answers for the student LM. GPT-3.5-Turbo and Gemini-1.0-Pro are required by their official APIs. Mixtral-8x7B-Instruct-v0.1 is required by the API hosted on Deepinfra: \url{https://deepinfra.com/mistralai/Mixtral-8x7B-Instruct-v0.1}.
\begin{table}[ht]
\centering
{\begin{tabular}{lc}
\toprule Parameter & Value \\ \midrule
Temperature & 0.8 \\
Max tokens & 512\\
Top p & 1\\
Presence penalty & 0\\
Frequency penalty & 0\\ \bottomrule
\end{tabular}}
\caption{\label{table_teacher_parse} Teacher LLMs parameter settings.}
\end{table}

\subsection{Student LM Parameters}
Experiments are performed with the Huggingface Trainer framework and Flash Attention~\cite{dao2022flashattention}. We use four Nvidia A100-80GB GPUs with FP16 for training and evaluation. The inference parameter settings across all datasets are shown in Table~\ref{table_student_parse}. The training hyperparameter settings across all datasets are shown in Table~\ref{table_training_hyper}. 

\clearpage

\begin{table}[h]
\centering
{\begin{tabular}{lc}
\toprule Parameter & Value \\ \midrule
Temperature & 0.3 \\
Max new tokens & 512\\
Top p & 0.9\\
Top k & 50\\
Do sample & True\\ \bottomrule
\end{tabular}}
\caption{\label{table_student_parse} Student LM inference parameter settings.}
\end{table}

\begin{table}[h]
\centering
{\begin{tabular}{lc}
\toprule Hyperparameter & Value \\ \midrule
Epoch & 10 \\
Batch size & 8\\
Learning rate & 1e-5\\
Warmup ratio & 0.03\\
Max seq length & 512\\
Optimizer & AdamW\\
Gradient accumulation steps & 2\\
Max grad norm & 0.3\\ \bottomrule
\end{tabular}}
\caption{\label{table_training_hyper} Student LM training hyperparameter settings.}
\end{table}

\section{Hyperparameter Tuning}
\label{sec:appendix_tune}
We tuned {$\alpha$} on the full training set, ranging between [0, 1], and showed the performance on the final test set in Figure~\ref{fig_alpha}. Moreover, for learning rate, epoch, and other hyperparameters, we tuned different values on the entire training set and compared their performance on the entire test set, as shown in Table~\ref{table_dataset}, to prevent underfitting or overfitting.

\section{Peer-Review Examples}
\label{sec:appendix_pr}
Table~\ref{table_pr_eg} provides detailed examples of the peer-review process on GSM8K and StrategyQA. It highlights instances where the causality between the teacher LLM's rationale and the final answer may be insufficient, and demonstrates how our peer-review mechanism effectively identifies the most confident rationales.

\section{Instruction Tuning Templates}
\label{sec:appendixB}
\begin{itemize}

\item Instruction tuning templates for learning from mistakes.
\begin{itemize}
\item For all benchmarks:  \\
``\#\#\# Instruction: Imagine you are a teacher, I will give you one student's incorrect answer to a question. You should point out the mistakes in the student's answer. \\
\#\#\# Input: \{\} \\
\#\#\# Response: \{\}'' \\
\end{itemize}
\vspace{-0.5cm}

\item Instruction tuning templates for learning from rationale.
\begin{itemize}
\item For benchmarks GSM8K and SVAMP:  \\
``\#\#\# Instruction: Answer the following question. Let's think step by step. \\
\#\#\# Input: \{\} \\
\#\#\# Response: \{\}'' \\
\item For benchmark strategyQA:  \\
``\#\#\# Instruction: Answer the following question. Let's think step by step. First, you should answer ``true'' or ``false''. Then, you should explain how you draw this conclusion. \\
\#\#\# Input: \{\} \\
\#\#\# Response: \{\}'' \\
\item For benchmark logiQA:  \\
``\#\#\# Instruction: Answer the following question based on the given context, query, and options. Let's think step by step.\\
\#\#\# Input: \{\} \\
\#\#\# Response: \{\}'' \\
\end{itemize}

\end{itemize}

\section{The Performance of Peer-Review between Two Teacher LLMs}
\label{sec:appendix_2teachers}
To explore the cooperation between teacher LLMs further, we conduct experiments on the same student model Llama2-7B-chat based on combinations of two different teacher LLMs. The results are shown in Table~\ref{table_2teachers}. It is found that the performance improvement still correlates to the teacher LLMs' abilities on benchmarks. However, the performance of combinations for two teacher LLMs lags behind the three-teacher distillation, which proves the necessity of choosing three teacher LLMs as reviewers.
\begin{table}[htb]
\centering
\resizebox{1\columnwidth}{!}{\begin{tabular}{lllll}
\toprule & \multicolumn{2}{c}{\textbf{Mathematical}} & \textbf{Commonsense} & \textbf{Logical}\\
Student LMs & GSM8K & SVAMP & StrategyQA & LogiQA \\ \midrule
Llama2-7B-chat & 15.62 & 39.67 & 47.02 & 18.74 \\
+Mixtral, Gemini & 27.37 & 49.67 & 63.32 & 34.56 \\
+Mixtral, GPT & 29.11 & 52.67 & 64.92 & 33.95 \\
+Gemini, GPT  & 32.37 & 55.00 & 62.01 & 34.25 \\
+Multiple  & \textbf{36.24} & \textbf{59.50} & \textbf{67.69} & \textbf{36.25} \\ \bottomrule
\end{tabular}}
\caption{\label{table_2teachers} Results of peer-review between two teacher LLMs.}
\vspace{-0.5cm}
\end{table}

\section{Case Study of Distillation Impact on Student LM's Output}
\label{sec:appendix_output}
Table~\ref{table_output_eg} provides the comparisons of student LM's behavior difference before and after the instruction tuning across four benchmarks.

\section{The Performance of Out-of-Distribution (OOD) Scenarios}
\label{sec:appendix_ood}
To evaluate the generalization abilities of different methods on out-of-distribution (OOD) data, we conducted experiments using one mathematical reasoning dataset as the training set and another dataset as the test set. Table~\ref{table_ood} highlights the performance of the FAIR method on Llama2-7B-chat in OOD scenarios.  

The results indicate a decrease in performance improvement compared to the original in-distribution scenarios in Table~\ref{table_results}. Specifically, the accuracy gains on GSM8K were smaller than those on SVAMP, likely due to the greater complexity of GSM8K. Despite this, our multiple-teacher distillation approach consistently outperforms all single-teacher methods under OOD conditions, demonstrating its robustness and generalizability.

\begin{table}[htb]
\centering
\resizebox{1\columnwidth}{!}{\begin{tabular}{lll}
\toprule & \multicolumn{2}{c}{\textbf{Mathematical}}\\
Student LMs & GSM8K & SVAMP \\ \midrule
Llama2-7B-chat+Mixtral  & 8.72 & 36.67 \\
Llama2-7B-chat+Gemini  & 10.54 & 46.00 \\
Llama2-7B-chat+GPT  & 12.36 & 47.67 \\
Llama2-7B-chat+Multiple  & \textbf{13.43} & \textbf{55.00} \\ \bottomrule
\end{tabular}}
\caption{\label{table_ood}The performance of FAIR on Llama2-7B-chat in out-of-distribution (OOD) scenarios. Specifically, we conducted experiments by training on SVAMP and testing on GSM8K, as well as training on GSM8K and testing on SVAMP.}
\end{table}

\begin{table*}[ht]
\centering
\small
\begin{tabular}{p{3cm}|p{13cm}}
\multicolumn{2}{c}{\textbf{GSM8K Example}}\\ \toprule
Question & ``Josh is saving up for a box of cookies. To raise the money, he is going to make bracelets and sell them. It costs \$1 for supplies for each bracelet and he sells each one for \$1.5. If he makes 12 bracelets and after buying the cookies still has \$3, how much did the box of cookies cost?'' \\ \midrule
Gold answer & ``He makes \$.5 profit on each bracelet because 1.5 - 1 = .5 He earns \$6 because 12 x .5 = 6 The cookies cost \$3 because 6 - 3 = 3 \#\#\#\# \textbf{3}'' \\ \midrule
GPT’s answer & ``Let's first calculate Josh's profit per bracelet: Profit per bracelet = Selling price - Cost price Profit per bracelet = \$1.5 - \$1 Profit per bracelet = \$0.5 Now, let's calculate Josh's total profit from making and selling 12 bracelets: Total profit = Profit per bracelet * Number of bracelets Total profit = \$0.5 * 12 Total profit = \$6 We know that after buying the cookies, Josh still has \$3 left. This means he spent \$6 - \$3 = \$3 on the box of cookies. Therefore, the box of cookies cost \textbf{\$3}.''	
\newline \textbf{Gemini’s score: 5, Mixtral’s score: 5} \\ \midrule
Gemini’s answer & ``\textbf{15}''
\newline \textbf{GPT’s score: 1, Mixtral’s score: 2} \\ \midrule
Mixtral’s answer & ``The box of cookies cost \$13. Here's the reasoning: 1. Josh spends \$1 on supplies for each bracelet, and he makes 12 bracelets. So, the total cost of supplies is 12 × \$1 = \$12. 2. He sells each bracelet for \$1.5, and he makes 12 bracelets. So, the total revenue from selling the bracelets is 12 × \$1.5 = \$18. 3. After buying the cookies, he still has \$3 left. This means that his total revenue (\$18) minus the cost of the cookies (\$X) equals the amount of money he has left (\$3). In other words, \$18 - \$X = \$3. 4. Solving for the cost of the cookies (\$X), we get \$X = \$18 - \$3 = \$15. However, the problem states that after buying the cookies and still having \$3, the cost of the cookies should be \$13, not \$15. \textbf{This might be a mistake in the question itself}.''	
\newline \textbf{GPT’s score: 2, Gemini’s score: 2} \\ \midrule
Peer-reviewed answers & \textbf{GPT’s answer} \\ \bottomrule
\end{tabular}

\vspace{2em}

\begin{tabular}{p{3cm}|p{13cm}}
\multicolumn{2}{c}{\textbf{StrategyQA Example}}\\ \toprule
Question & ``Would 2020 Toyota Supra lag behind at a Nascar rally?'' \\ \midrule
Gold answer & ``\textbf{True}. The 2020 Toyota Supra has a top speed of 155 MPH. Nascar stock cars routinely exceed 200 MPH.'' \\ \midrule
GPT’s answer & ``\textbf{False}. Explanation: 1. The 2020 Toyota Supra is not a car typically used in NASCAR racing. NASCAR races primarily feature specially designed stock cars that meet specific performance and safety regulations. 2. The 2020 Toyota Supra is a production sports car designed for road use, not for NASCAR track racing. While it may have impressive performance capabilities on the road, it is not optimized for NASCAR oval track racing conditions. 3. NASCAR races involve high speeds, close competition, and specific handling characteristics that are different from what the 2020 Toyota Supra is designed for. \textbf{The Supra would likely struggle to keep up with dedicated NASCAR race cars} in a NASCAR rally due to these differences in design and performance characteristics.''	
\newline \textbf{Gemini’s score: 4, Mixtral’s score: 2} \\ \midrule
Gemini’s answer & ``\textbf{True}. The 2020 Toyota Supra is a sports car, while NASCAR vehicles are purpose-built race cars. The Supra would not be competitive in a NASCAR race because it is not designed for that type of racing. NASCAR vehicles are much more powerful and have a much better suspension system than the Supra. The Supra would also be at a disadvantage because it is not as aerodynamic as a NASCAR vehicle.''	
\newline \textbf{GPT’s score: 4, Mixtral’s score: 5} \\ \midrule
Mixtral’s answer & ``\textbf{True}. NASCAR race cars are designed for high-speed oval track racing, with powerful V8 engines, aerodynamic bodies, and specialized suspension systems. The Supra, on the other hand, is a street-legal sports car, with a turbocharged inline-six engine and a more versatile suspension setup. If the Supra rebuilds the engine and improve other systems, \textbf{it is likely suited for the unique demands of NASCAR racing and keep up with NASCAR vehicles.}''	
\newline \textbf{GPT’s score: 2, Gemini’s score: 3} \\ \midrule
Peer-reviewed answers & \textbf{Gemini’s answer} \\ \bottomrule
\end{tabular}
\caption{\label{table_pr_eg} Detailed examples of peer-review process on different benchmarks}
\end{table*}

\begin{table*}[ht]
\centering
\small
\begin{tabular}{p{4cm}|p{12cm}}
\multicolumn{2}{c}{\textbf{GSM8K Example}}\\ \hline
Question & Weng earns \$12 an hour for babysitting. Yesterday, she just did 50 minutes of babysitting. How much did she earn?   \textbf{Gold answer: }\textcolor{ForestGreen}{\$10} \\ \hline
Student’s output before tuning & Answer: \$18.  Explanation:Weng earns \$12 per hour for babysitting. Since she did 50 minutes of babysitting, she earned \$12 x 0.75 = \$9 per 50 minutes. So, she earned \textcolor{Red}{\$9 x 2 = \$18}. \\ \hline
Teachers' correct rationale & Weng earned \$10 for 50 minutes of babysitting. Here's the logic behind this: 1. We know that Weng earns \$12 per hour. 2. To find out how much she earns in 50 minutes, we first need to convert the time into hours. 3. We know that 1 hour equals 60 minutes, so 50 minutes is equal to 50/60 hours, which simplifies to 0.833 hours. 4. Now, we can calculate how much Weng earns for 0.833 hours by multiplying her hourly rate by the number of hours: \textcolor{ForestGreen}{\$12 x 0.833 = \$10}. So, Weng earned \$10 for 50 minutes of babysitting. \\ \hline
Teachers' mistakes feedback & The student's answer contains a mistake in calculating how much Weng earned for 50 minutes of babysitting. The error is in multiplying the hourly rate by 0.75 to get the amount earned in 50 minutes. Instead, the correct approach would be to calculate the rate per minute by dividing the hourly rate by 60, and then multiplying that rate by the number of minutes worked. The correct answer is \$10, not \$18. \\ \hline
Student’s output after tuning & Weng earns \$12 per hour, so she earns \$12 / 60 = \$0.20 per minute.
Since she did 50 minutes of babysitting, she earned \textcolor{ForestGreen}{\$0.20 x 50 = \$10}. \\ \hline
\end{tabular}

\vspace{0em}

\begin{tabular}{p{4cm}|p{12cm}}
\multicolumn{2}{c}{\textbf{SVAMP Example}}\\ \hline
Question & Paul got a box of 110 crayons for his birthday. During the school year he gave 90 crayons to his friends while he lost 412 crayons. How many more crayons did he lose than those he gave to his friends? \textbf{Gold answer: }\textcolor{ForestGreen}{322} \\ \hline
Student’s output before tuning & Answer: 392. Explanation: Paul gave 90 crayons to his friends, so he had 110 - 90 = 20 crayons left. So he lost \textcolor{Red}{412 - 20 = 392}. \\ \hline
Teachers' correct rationale & Paul lost 412 crayons and gave away 90 crayons. To find out how many more crayons he lost than those he gave to his friends, we can subtract the number of crayons he gave away from the number of crayons he lost: \textcolor{ForestGreen}{412 - 90 = 322} Therefore, Paul lost 322 more crayons than those he gave to his friends. \\ \hline
Teachers' mistakes feedback & The mistakes in the student's answer are as follows: The student calculated the number of crayons Paul lost by subtracting the remaining crayons (20) from the total number of crayons he lost (412), which is incorrect. The correct calculation should be 412 - 90 = 322. Therefore, the correct answer should be 412 - 90 = 322, indicating that Paul lost 322 more crayons than he gave to his friends. \\ \hline
Student’s output after tuning & Paul lost \textcolor{ForestGreen}{412 crayons - 90 crayons = 322} crayons more than those he gave to his friends. \\ \hline
\end{tabular}

\vspace{0em}

\begin{tabular}{p{4cm}|p{12cm}}
\multicolumn{2}{c}{\textbf{StrategyQA Example}}\\ \hline
Question & Would someone in Mumbai refer to Solanum melongena as an eggplant? \textbf{Gold answer: }\textcolor{ForestGreen}{False} \\ \hline
Student’s output before tuning & \textcolor{Red}{True}: In Mumbai, Solanum melongena is commonly referred to as an eggplant. \\ \hline
Teachers' correct rationale & \textcolor{ForestGreen}{False}. In Mumbai, people would more commonly refer to ``Solanum melongena'' as ``brinjal'' or ``baingan'' rather than ``eggplant.''  While ``eggplant'' is widely used in American English, in India, especially in regions like Mumbai, the term ``brinjal'' is more common. ``Baingan'' is a Hindi word used across many parts of India. \\ \hline
Teachers' mistakes feedback & The student's answer contains a mistake in the explanation provided. The correct answer to the question is False, but the student mistakenly provided a hint that supports the correct answer. The correct information should be that in Mumbai, Solanum melongena is commonly referred to as a brinjal, not an eggplant. \\ \hline
Student’s output after tuning & \textcolor{ForestGreen}{Answer: False}.  Solanum melongena is not referred to as ``eggplant'' in Mumbai. The term ``eggplant'' is specifically used in American English and Solanum melongena is known as ``brinjal'' in Mumbai. So, the answer is false. \\ \hline
\end{tabular}

\vspace{0em}

\begin{tabular}{p{4cm}|p{12cm}}
\multicolumn{2}{c}{\textbf{LogiQA Example}}\\ \hline
Question & Some Cantonese don't like chili, so some southerners don't like chili. Which of the following can guarantee the above argument? Options: [``Some Cantonese love chili.'', ``Some people who like peppers are southerners.'', ``All Cantonese are southerners.'', ``Some Cantonese like neither peppers nor sweets.''].   \textbf{Gold answer: }\textcolor{ForestGreen}{``All Cantonese are southerners.''} \\ \hline
Student’s output before tuning & \textcolor{Red}{Answer: [``Some Cantonese love chili.'']} Rationale: The argument states that ``Some Cantonese don't like chili,'' which implies that not all Cantonese people dislike chili. Therefore, the statement ``Some Cantonese love chili'' is the only option that can guarantee the argument. \\ \hline
Teachers' correct rationale & \textcolor{ForestGreen}{The correct option is: ``All Cantonese are southerners.''}   Explanation: The argument states that some Cantonese don't like chili, and to conclude from this that some southerners don't like chili, we need to ensure that Cantonese are included within the broader category of southerners. If all Cantonese are southerners, then the statement about ``some Cantonese not liking chili'' can be applied to southerners as well, making the argument valid. \\ \hline
Teachers' mistakes feedback & The student's answer is incorrect. The statement ``Some Cantonese love chili'' does not guarantee the argument about ``southerners'' provided in the question. The correct answer should be ``All Cantonese are southerners.'' This option directly connects the dislike for chili among some Cantonese to the fact that some southerners don't like chili, making it the most relevant choice to support the argument. \\ \hline
Student’s output after tuning & \textcolor{ForestGreen}{Answer: [``All Cantonese are southerners.'']}  Explanation: The argument states: ``Some Cantonese don't like chili, so some southerners don't like chili.'' To guarantee this argument, we need to establish that some southerners (specifically Cantonese) don't like chili.
 \\ \hline
\end{tabular}

\caption{\label{table_output_eg} Case study of distillation impact on student LM's output}
\end{table*}

\end{document}